\documentclass[lettersize,journal]{IEEEtran}
\usepackage{amsmath,amsfonts}
\usepackage{algorithmic}
\usepackage{algorithm}
\usepackage{array}
\usepackage[caption=false,font=normalsize,labelfont=sf,textfont=sf]{subfig}
\usepackage{textcomp}
\usepackage{xcolor}
\usepackage{stfloats}
\usepackage{url}
\usepackage{verbatim}
\usepackage{graphicx}
\usepackage{cite}
\usepackage{multicol}
\usepackage{xcolor,colortbl}
\usepackage[hidelinks]{hyperref}
\usepackage{scalerel}
\usepackage{tikz}
\usetikzlibrary{svg.path}

\definecolor{orcidlogocol}{HTML}{A6CE39}
\tikzset{
  orcidlogo/.pic={
    \fill[orcidlogocol] svg{M256,128c0,70.7-57.3,128-128,128C57.3,256,0,198.7,0,128C0,57.3,57.3,0,128,0C198.7,0,256,57.3,256,128z};
    \fill[white] svg{M86.3,186.2H70.9V79.1h15.4v48.4V186.2z}
                 svg{M108.9,79.1h41.6c39.6,0,57,28.3,57,53.6c0,27.5-21.5,53.6-56.8,53.6h-41.8V79.1z M124.3,172.4h24.5c34.9,0,42.9-26.5,42.9-39.7c0-21.5-13.7-39.7-43.7-39.7h-23.7V172.4z}
                 svg{M88.7,56.8c0,5.5-4.5,10.1-10.1,10.1c-5.6,0-10.1-4.6-10.1-10.1c0-5.6,4.5-10.1,10.1-10.1C84.2,46.7,88.7,51.3,88.7,56.8z};
  }
}

\newcommand\orcidicon[1]{\href{https://orcid.org/#1}{\mbox{\scalerel*{
\begin{tikzpicture}[yscale=-1,transform shape]
\pic{orcidlogo};
\end{tikzpicture}
}{|}}}}

\usepackage{hyperref} 
\newcommand{\email}[1]{\href{mailto:#1}{\mbox{#1}}}

\newcommand\copyrighttext{%
  \footnotesize \textcopyright 2023 IEEE. Paper published in IEEE Trans. Comput. Imaging. Personal use of this material is permitted.
  Permission from IEEE must be obtained for all other uses, in any current or future
  media, including reuse of any copyrighted component of this work in other works.
  DOI: \href{https://doi.org/10.1109/TCI.2023.3288335}{10.1109/TCI.2023.3288335}}
\newcommand\copyrightnotice{%
\begin{tikzpicture}[remember picture,overlay]
\node[anchor=south,yshift=10pt] at (current page.south) {\fbox{\parbox{\dimexpr\textwidth-\fboxsep-\fboxrule\relax}{\copyrighttext}}};
\end{tikzpicture}%
}

\begin{document}

\title{Depth Estimation and Image Restoration by Deep Learning from Defocused Images}

 \author{Saqib~Nazir \orcidicon{0000-0003-0098-1868}\ ,~\IEEEmembership{Student Member,~IEEE,}
      Lorenzo~Vaquero \orcidicon{0000-0002-1874-3078},
      Manuel~Mucientes \orcidicon{0000-0003-1735-3585},\\
      V\'ictor~M. Brea \orcidicon{0000-0003-0078-0425},
      and~Daniela Coltuc \orcidicon{0000-0002-0237-7878}\ 
 \thanks{This project has received funding from the European Union’s Horizon $2020$ research and innovation programme under the Marie Skłodowska-Curie grant agreement No. $860370$. The last author acknowledges financial support from UEFISCDI Romania grant $31/01.01.2021$ PN III, 3.6 Support. This research was also partially funded by the Spanish Ministerio de Ciencia e Innovación [grant number PID2020-112623GB-I00], and the Galician Consellería de Cultura, Educación e Universidade [grant numbers
ED431C 2018/29, ED431C 2021/048, ED431G 2019/04]. These grants are co-funded by the European Regional Development Fund (ERDF). Lorenzo Vaquero is supported by the Spanish Ministerio de Universidades under the FPU national plan (FPU18/03174).}
  \thanks{S. Nazir and D. Coltuc are with CEOSpaceTech, University POLITEHNICA of Bucharest (UPB), Bucharest, Romania (e-mail: \email{saqib.nazir@upb.ro} ; \email{daniela.coltuc@upb.ro}).}
  \thanks{L. Vaquero, M. Mucientes, and V. Brea are with Centro Singular de Investigaci\'on en Tecnolox\'ias Intelixentes (CiTIUS) University of Santiago de Compostela (USC), Santiago de Compostela, Spain (e-mail: \email{lorenzo.vaquero.otal@usc.es};  \email{manuel.mucientes@usc.es} ; \email{victor.brea@usc.es}).}
  }

\maketitle
\copyrightnotice

\begin{abstract}
Monocular depth estimation and image deblurring are two fundamental tasks in computer vision, given their crucial role in understanding 3D scenes. Performing any of them by relying on a single image is an ill-posed problem. The recent advances in the field of Deep Convolutional Neural Networks (DNNs) have revolutionized many tasks in computer vision, including depth estimation and image deblurring. When it comes to using defocused images, the depth estimation and the recovery of the All-in-Focus (Aif) image become related problems due to defocus physics. Despite this, most of the existing models treat them separately. There are, however, recent models that solve these problems simultaneously by concatenating two networks in a sequence to first estimate the depth or defocus map and then reconstruct the focused image based on it. We propose a DNN that solves the depth estimation and image deblurring in parallel.  Our Two-headed Depth Estimation and Deblurring Network (2HDED:NET) extends a conventional Depth from Defocus (DFD) networks with a deblurring branch that shares the same encoder as the depth branch.  The proposed method has been successfully tested on two benchmarks, one for indoor and the other for outdoor scenes: NYU-v2 and Make3D. Extensive experiments with 2HDED:NET on these benchmarks have demonstrated superior or close performances to those of the state-of-the-art models for depth estimation and image deblurring.
\end{abstract}


\section{Introduction}
\IEEEPARstart{D}{epth} estimation from a single image is a key problem in computer vision, where it spans a lot of applications. 
Robotics, augmented reality, human-computer interaction or computational photography, to give only several examples, benefit from depth estimation. With the recent advancements in 3D computer vision and the newly emerging tasks like semantic segmentation or 3D object detection, depth estimation has become even more important. 

The depth can be measured by specialized devices or can be inferred from images and videos. For outdoor scenes, LIDAR or stereo systems are typically used to measure the depth of the scene. For indoor scenes, Time of Flight (ToF) cameras like RGBD Kinect from Microsoft, is used to capture depth information in addition to the RGB images. However, the applicability of these devices is limited. ToF cameras are not working properly in the outdoors, being limited to $30$m at best, while the LIDAR may produce poor-quality depth maps because of infrared interference. 
These physical limitations, the sparse nature of the measurements, and the cost of the devices have fostered the research in the direction of obtaining depth from  images or videos taken with commercial cameras. 
Here, although the performance of depth estimation methods is steadily increasing, there are still major problems related to the accuracy and resolution of the estimated depth maps.   
\begin{figure}[!t]
\centering
\includegraphics[width=1\linewidth]{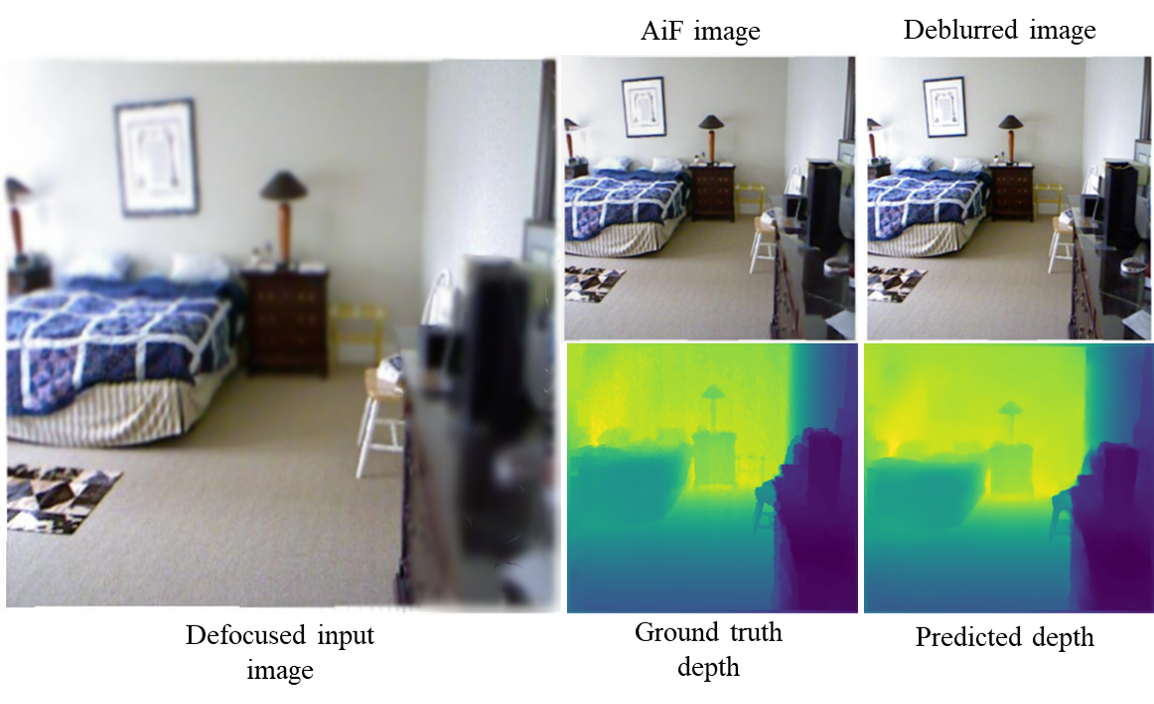}
\caption{Example of image deblurring and depth prediction using 2HDED:NET in a scene from the NYU-v2 dataset.
\textcolor{black}{AiF is an all-in-focus image that serves as ground truth for deblurring. The AiF and ground-truth depth are captured by an RGBD Kinect camera.}  }
\label{fig_1}
\end{figure}

Image deblurring is a classical problem in low-level computer vision, and a prepossessing step in numerous applications such as face detection, classification, object recognition, or misfocus correction. Object motion, camera shake, or out-of-focus are common causes for the blur appearing in the images taken with a camera. The goal of image deblurring is to recover an AiF image with all the details and sharp edges from its defocused counterpart. 

The main objective of the network proposed in this paper is to estimate the depth and remove blur from a single out-of-focus image. Figure~\ref{fig_1} shows an example of Depth from Defocus (DFD) and image deblurring. In this example, the network estimates a dense depth map and reconstructs the AiF image from a defocused image.

Most of the DNNs dedicated to depth estimation work on AiF images \cite{Godard_un, saxena2008make3d, eigen2014depth}. The exploitable information  in such images is limited to the scene geometry, which explains the lower performances compared to LIDAR or ToF cameras.  
The defocus blur is a complementary cue that can help to improve the depth accuracy.

DFD has been widely investigated in the past \cite{tang2015depth}. 
The first DFD  methods were focused on the depth related to the blur amount and as a result, they suffered from insensitivity in the Depth of Field (DoF) region and uncertainty regarding the object position with respect to the in-focus plane. \textcolor{black}{The use of coded apertures \cite{levin2007image, masoudifar2022depth}, dual images or focal stacks \cite{hazirbas2018deep, lien2020ranging, huang2020light} has alleviated such problems.}

In many applications, including DFD \cite{Anwar, Carvalho, Fu, Song}, DNN models outperform the classical methods, due to the ability of learning more complex features.
The features learned from defocused images combine both the scene geometry and the blur for more accurate depth estimation.
In the last years, a series of DNNs has been proposed for image deblurring as well \cite{AIFNet,Anwar,DPD,zhang2022deep, li2022survey}. 

Although the depth and defocus blur are closely related, the deblurring and depth estimation have been generally, treated as separate problems by deep learning. There are however some rare exceptions like the method of Anwar et al. \cite{Anwar}, which concatenates two networks to first estimate the depth map and then, based on this depth map, restores a focused image by pixel-wise non-blind deconvolution. 
{\color{black}More precisely, in \cite{Anwar} a fully convolutional neural network with 13 layers provides a pixel level feature map, then a patch pooling layer turns the patches around predefined key points into fixed size feature map, which are further propagated through a shallow fully connected network to estimate a dense depth map. The deblurring is done by deconvolution with kernels calculated for every pixel of the RGB image, by using the estimated depth.}

\begin{figure*}[t]
\centering{\includegraphics[height=10cm, width=14cm]{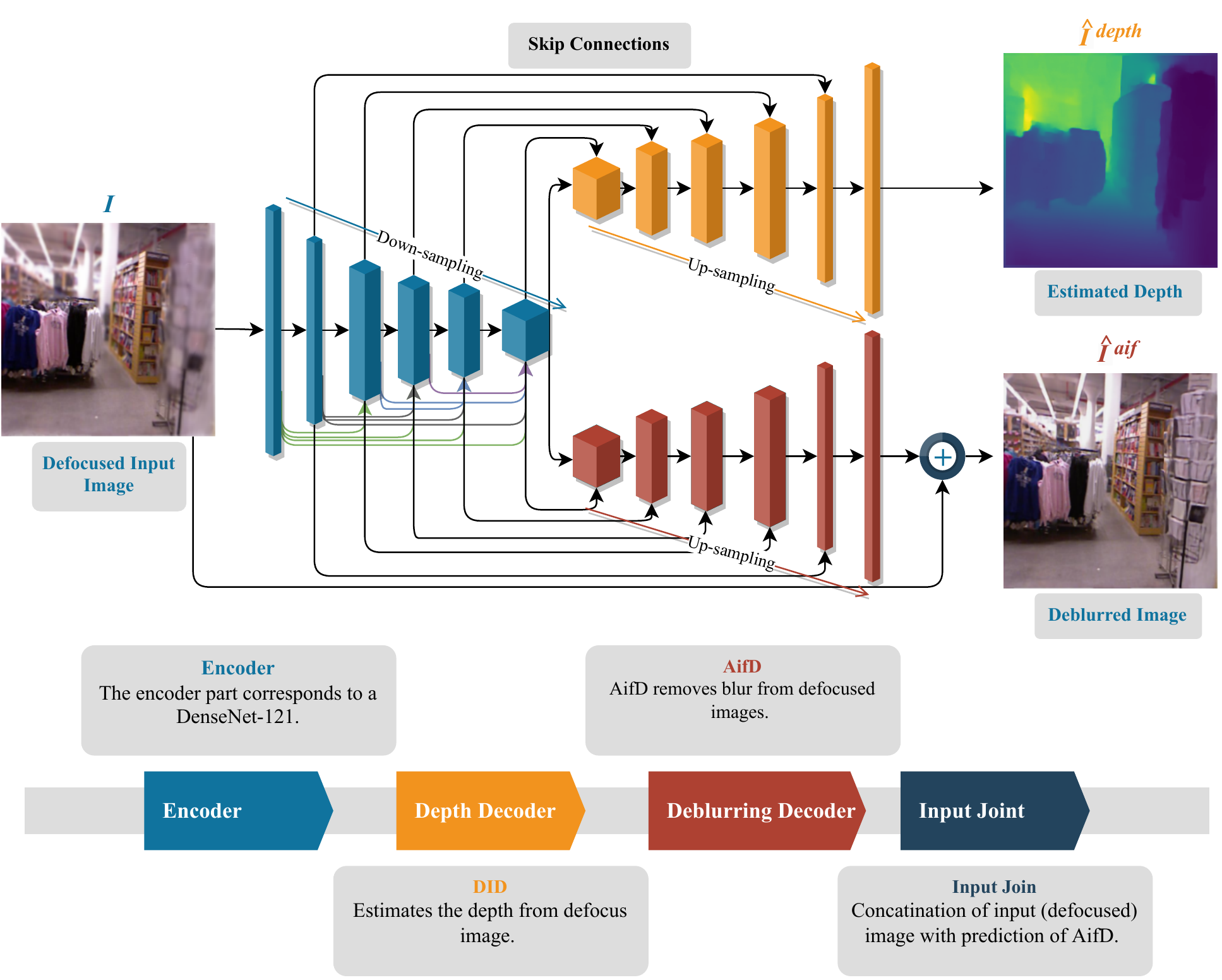}}
\caption{2HDED:NET architecture consists of one encoder and two decoders that work in parallel. The upper Head estimates the depth map and the lower one the AiF image. The network is fed in with defocused RGB images.}
\label{fig:archi}
\end{figure*}

\textcolor{black}{It is known that architectures with independent and task-dedicated branches and their loss terms combined decrease overfitting in the training phase, and permit to execute any of the tasks in the inference time. In this line, we propose a DNN that solves the problem of DFD and image deblurring in parallel. }  
The proposed two-headed network, called 2HDED:NET, estimates the depth and deblurs the image in a balanced way by giving the same importance to both tasks. 
The network consists of three modules: \romannumeral 1) an encoder for multi-level feature extraction from the defocused image, 
\romannumeral 2) a depth estimation decoder (DED) for the DFD, \romannumeral 3) an AiF decoder (AifD) for image deblurring (Fig.~\ref{fig:archi}). The heads interact with each other during training, allowing the encoder to learn semantically rich features that are well-suited for both tasks. 

\textcolor{black}{Unlike Anwar et al. in \cite{Anwar}, where the deblurring depends on the intermediate result of depth estimation, our 2HDED:NET generates an AiF image, which is not any more constrained by depth accuracy. 
Separating the deblurring and depth estimation branches also makes AifD self-sufficient and better able to perform the deblurring task without relying on an estimated depth map.
}

\textcolor{black}{2HDED:NET is a typical Multitask Learning (MTL) neural network with hard sharing of parameters. The encoder layers are shared by both depth estimation and deblurring tasks while the two decoders remain task-specific. 
Comparing to the single task networks, the MTL networks benefit from a series of advantages:  
an augmented training set, relevant feature learning by attention focusing, easier learning of features from less complex models, reduced risk of overfitting, and better generalization to new tasks [21]. The foundations of MTL by hard parameter sharing had been laid by Caruana in 1997 \cite{caruana1997multitask}, and two recent surveys of MTL can be found in \cite{ruder2017overview, crawshaw2020multi}. The MTL technique has been used successfully in computer vision applications as well as in other areas such as natural language processing and drug discovery. Two recent applications closely related to our application are addressed in \cite{lu2020multi,wang2022semi}, where the depth map and semantic segmentation are learned by MTL.}

The architecture of 2HDED:NET is straightforward, simple, and easy to train. 
With its double functionality -- depth estimation and deblurring -- 2HDED:NET emulates a Kinect-type camera on a commercial camera with limited DoF.
A special feature of 2HDED:NET is that after training, the depth estimation head is no longer necessary to recover a sharp image and vice versa.

We define a hybrid loss function to train 2HDED:NET. It embeds specific cost functions for depth and deblurring like $L1$ norm and Charbonier loss \cite{Carvalho, xu2021edpn}, as well as specific regularizations like gradient-based smoothing \cite{Xian} and maximization of Structural Similarity Index Measure (SSIM).

We run extensive experiments on the NYU-v2 and Maked3D benchmarks in order to evaluate the performance of the 2HDED:NET and to compare with the state-of-the-art methods for depth estimation  and image deblurring. In most cases, 2HDED:NET generates better results. For training, 2HDED:NET uses two types of ground truth, the depth and the AiF images in the benchmarks. As input, synthetically defocused images are generated using the thin lens model. Hence, the prior information is consistent with that of any other network for depth estimation. The supplement includes the mathematical model of the defocus blur, which is proven effective through our results.

\par
The main scientific contributions of our work are:
\begin{itemize}
  \item A parallel architecture, namely 2HDED:NET, that enables the recovery of AiF images and the generation of depth maps from a single defocused image.
  \item The architecture has the merit to achieve a balanced generation of both depth maps and AiF images while assigning equal significance to both tasks.
  \item A hybrid loss function that combines losses and regularizations from both depth estimation and deblurring and enforces the encoder to learn much richer semantic features.  
  \item Extensive experimental results on NYU-v2 and Make3D datasets enriched with synthetic defocused images, confirm the effectiveness of our approach.
\end{itemize} \par
The remainder of this paper is organized as follows. In Section II, we provide an overview of the related work. In Section III, we present our methodology. The experimental setup and results are reported in Sections IV,  and V, and, finally, in Section VI, we present  conclusions and future work. 

\section{Related work}

This section briefly reviews the state-of-the-art DNN-based solutions for DFD and deblurring.

\subsection{Single image depth estimation}
The success of DNN models in various fields of computer vision, such as image segmentation and classification, has prompted the scientific community to consider using DNNs for depth estimation as well. Saxena et al. \cite{saxena2008make3d} presented one of the first solutions for monocular depth estimation with deep learning methods. They estimate the depth with a multi-scale architecture and the Markov Random Field (MRF). Eigen et al. \cite{eigen2014depth} presented one of the most successful works by developing a multi-scale architecture to extract information from a scene at global and local levels in order to estimate the depth map. Laina et al. \cite{laina} proposed an encoder-decoder network with a fast up-projection block. Cao et al.\cite{cao2017estimating} relied on Conditional Random Fields (CRF) to improve the accuracy of the depth maps. GANs have also been utilized for depth prediction. Jung et al. in \cite{jung} and Carvalho et al. in \cite{Carvalho2} implemented an adversarial loss for depth prediction.  
Most of the evoked networks solve the problem of monocular depth estimation by using in-focus images as input and ignoring the defocus blur, which is however an important cue in depth estimation. 
\subsection{Depth from defocus}
Defocus blur occurs when images are captured with limited DoF. All cameras have a limited DoF, which is controlled by the camera aperture diameter. Although the blur exists also in this range, it is not perceived by the human eye. 

To estimate depth from a single defocused image, Carvalho et al. \cite{Carvalho}  built on a dense network DenseNet-121 with skip connections that improved the state-of-the-art results of the time. To handle the DFD problem, Gur et al.  \cite{Gur} designed a convolution layer based on the Point Spread Function (PSF) to train an unsupervised network. Anwar et al. \cite{Anwar} trained a fully connected cascaded deep neural network inspired by the VGG-16 model on dense overlapping fields to estimate depth from a single defocused image. In \cite{Fu}, Fu et al. proposed a multi-scale network structure to obtain high-resolution depth maps using spacing-increasing discretization and a simple regression loss.   

\subsection{Image deblurring}
 Blind image deblurring has been always a difficult problem. Since the advent of DNNs, several models were designed to reduce the blur from a single image. 
 The first ones directly remove the blur, such as Nah et al. \cite{nah2017deep}, who used a multiscale loss function to train their model. 
 Tao et al. \cite{tao2018scale} improved their work by using joint network parameters at different scales. Kupyn et al. proposed DeblurGAN to reconstruct AiF images from defocused images by using an adversarial loss function \cite{kupyn2018deblurgan}. 

The strategy adopted in more recent papers is to first estimate a defocus/depth map and later use this information for image deblurring.    
Thus, Lee et al. \cite{Lee} introduced a deep architecture along with a domain-matching approach to estimate the defocus map of an image, and also presented a large dataset for training DNNs.
Very recently, by estimating defocus maps, works like \cite{AIFNet, Anwar} use the amount of blur per pixel to reconstruct the entire deblurred image.

{\color{black}Zhang Kai et al. proposed two general methods for image restoration, with deblurring as a particular case in \cite{zhang2017beyond,zhang2017learning}. In \cite{zhang2017beyond}, it is shown that by separating the fidelity from the regularization term in the energy function, the optimization problem can be solved by plugging a denoising neural  network in a  Half Quadrating Splitting framework. The method is tested for denoising, super-resolution and deblurring. In \cite{zhang2017learning}, the authors proposed a convolutional neural network for blind Gaussian denoising. The network removes the latent clean image and estimates the residual Gaussian noise with unknown level. By observing that the image degradation model for Gaussian denoising can be converted to other restoration problems, the authors successfully apply it to image super-resolution and JPEG deblocking. 

The priors on image model play an important role in image restoration by optimization. Zha et al. proposed in \cite{zha2021triply} a low-rank and deep image model with three complementary priors: internal and external, shallow and deep, and non-local and local priors. The model is successfully tested on image deblurring, restoration after compressive sensing, and JPEG deblocking. The sole non-local self-similarity prior is used by Zha et al. in \cite{zha2020image} for image restoration by using the Expectation Maximization algorithm with image deblurring, denoising, and deblocking as applications. 
}

\subsection{Joint DFD and image deblurring}
The survey of the literature has revealed only two DNN models that addressed both depth estimation and blurring or deblurring process. Gur et al. proposed a network to estimate the depth from a single defocused image in \cite{Gur}. Unlike the supervised learning networks, they adopted a self-supervised learning approach with a loss function based on the difference between the defocused input image and a defocused image estimated by a second network. This second network implements the blur model and creates a synthetically defocused image by using the estimated depth.

The closest approach to our architecture is the model proposed by Anwar et al. in \cite{Anwar}. They train a cascade of two smaller networks to estimate a depth map, which is then used to compute kernels for restoring the AiF image by pixel-wise non-blind deconvolution.

\section{2HDED:NET Architecture}
\begin{table}[b]
\centering
\caption{Size of output features and input/output channels of each layer of 2HDED:NET.}
\vspace{2mm}
\begin{tabular}[ht]{|c|c|c|c|}
\hline
 {\textit{Layer}}  & {\textit{Output size}}  & \ {\textit{Input/C}}  & {\textit{Output/C}}  \ \\
\hline
 Conv1 & $128\times128$ &  $3$ & $64$ \\
 Conv2 & $64\times64$ & $64$  & $128$\\
 Conv3 & $32\times32$ &  $128$ & $256$ \\
 Conv4 & $16\times16$ & $256$ & $512$\\
 Conv5 & $8\times8$ & $512$ & $1024$\\
\hline
 Dconv5 & $8\times8$ & $1024$ & $1024$ \\
 Dconv4 & $16\times16$ & $1024$ & $512$ \\
 Dconv3 & $32\times32$ &  $512$ & $256$ \\
 Dconv2 & $64\times64$ & $256$  & $128$ \\
 Dconv1 & $128\times128$ &  $128$ & $64$ \\
     \hline
 Pred-depth & $256\times256$ &  $64$ & $1$ \\
\hline
 Pred-deblurring & $256\times256$ &  $64$ & $3$ \\
\hline
\end{tabular}
\label{tab:layers}
\end{table}
Figure~\ref{fig:archi} depicts the architecture of 2HDED:NET. Given a single defocused image $I$, the goal of our network is to estimate the depth map $\widehat{I}^{depth}$ and to restore the AiF image $\widehat{I}^{aif}$.
As shown in  Fig.~\ref{fig:archi}, 2HDED:NET consists of one encoder and two decoders that output the depth map and AiF image in parallel. By utilizing the features learned by the same encoder, both heads can mutually benefit from each other. 2HDED:NET is a supervised method that requires the ground truth depth as well as the AiF images for training. \par
\subsection{Encoder}
For the encoder network, we use the DenseNet-121 \cite{Densenet}. As its name suggests, DenseNet consists of densely connected layers. The main feature of DenseNet-121 is that this network reuses the features of each layer by concatenating them with the features of the next layer, rather than simply aggregating them like ResNet50. The goal of the concatenation is to use the features obtained in the previous layers in the deeper layers. This is referred to as "feature reusability". DenseNets can learn mappings with fewer parameters than a typical CNN since there are no redundant maps to learn. Similar to \cite{Carvalho}, we replace the max-pooling layer with a 4$\times$4 convolutional layer to reduce resolution while increasing the number of the feature channel maps. We use skip connections between the encoder and decoder parts to simplify learning. The skip connections prevent the problem of the gradient disappearing since the subsequent layers focus on solving residuals rather than completely new representations. The encoder helps to obtain multi-resolution features from the input image, which are useful for the two tasks that 2HDED:NET performs. Further information about the encoder's output size, input, and output channels can be seen in Table \ref{tab:layers}. 

\subsection{Depth Estimation Head}
The Depth Estimation Decoder (DED) is inspired by \cite{Carvalho}.
It consists of 5 decoding layers, each with 4$\times$4  
convolution that increases the resolution of the feature map, followed by a 3$\times$3 convolution that reduces the aliasing effect of upsampling. 
Batch normalization and ReLU functions are included after each convolutional layer to make learning more stable and to allow the representation of nonlinearities. Table \ref{tab:layers} shows how decoder layers upsample the input using transpose convolutions. The output of DED is one channel depth map. \par

\subsection{Deblurring Head}
We refer to the deblurring decoder as to AiF decoder (AifD). Unlike DED, the output of AifD is a three-channel RGB image. We use an input joint layer to aggregate the defocused input image with the output of AifD as in \cite{AIFNet, ED-DSRN} for the final prediction. The content of the defocused image and the corresponding prediction from AifD are embedded in the input joint layer, giving this head more detailed guidance for learning deblurring.
Unlike methods that use pipeline processing, where the depth or defocus map is first predicted and then the Aif image is recovered, our deblurring head is not based on such estimates, avoiding reliance on insufficient depth maps in some cases.  \par

An important feature of our solution is that once 2HDED:NET is fully trained, we are still able to perform a task when the other head is removed, e.g. we can perform DFD without the AifD head and vice versa.   \par

\subsection{Loss functions}
\label{lossfunctions}

The training of the 2HDED:NET is supervised simultaneously by ground truth depth maps and AiF images. To consider this dual information, we propose a loss function with two terms, one that accounts for the depth loss and another for the deblurred image. These two components are balanced to have approximately equal contributions.   

\subsubsection{Depth loss}
Most of the deep learning methods proposed for depth estimation have been trained with pixel-wise regression-based loss functions calculated as the mean of absolute differences ($L1$ norm), squared differences ($L2$ norm), or combinations of them \cite{Carvalho}.  
 
As the loss function for depth estimation, we resort to $L1$ norm, known for the ability to estimate sparse solutions as it is the case for depth maps  \cite{Carvalho,nazir2hded}: 
\begin{equation}
{L}_{1}^{Depth} = \frac{1}{n} \sum^n_{i=1} | {\widehat{I}}_{i}^{depth} - {I}_{i}^{depth}| \label{eq1}
\end{equation}
where ${\widehat{I}}^{depth}$ is the estimated depth, ${I}^{depth}$ the ground truth, $i$ is the current pixel and $n$ is the number of pixels.

Often, this loss is complemented by a smoothing regularization term that has the role of removing the low amplitude structures in the depth map while sharpening the main  edges \cite{HU, Godard, Gur, Xian}.
In the case of our network, we improve the depth accuracy by combining $L1$ norm with the smoothing term commonly used in supervised learning and defined as \cite{Xian}:
\begin{equation}
 {L}_{grad} = \frac{1}{n} \sum_{i}|\Delta_x R_i|+|\Delta_y R_i|
\label{eq2}
\end{equation}
where $R_i=\widehat{I}_{i}^{depth} - {I}_{i}^{depth}$ and $\Delta_x$ and $\Delta_y$ are the spatial derivatives with respect to the x-axis and y-axis. 
As a result, the overall depth loss function is defined as (3):
\begin{equation}
L_{depth} = {L}_{1}^{Depth} +\mu {L}_{grad} 
\label{depth_loss}
\end{equation}
where $\mu$ is a weighting coefficient set to 0.001. 

\subsubsection{Deblurring loss}
Various loss functions have been proposed to train the DNNs for image deblurring.
Pixel-wise content loss functions like $L1$ and $L2$ norm are the most common \cite{timofte2016seven, lim2017enhanced}. 

To train 2HDED:NET, we test $L1$ norm and Charbonnier loss function \cite{charbonnier}, which is the smoothed version of $L1$. 
Charbonnier loss is calculated as a squared error between the estimated deblurred image ${\widehat{I}}^{aif}$ and the ground truth AiF image ${I}^{aif}$:
\begin{equation}
{L}_{charb} = \frac{1}{n} \sum_{i=1}^{W} \sum_{j=1}^{H} \sqrt{({\widehat{I}}_{i,j}^{aif} - {I}_{i,j}^{aif})^2+\epsilon^2}  \label{eq3}
\end{equation}
where $\epsilon$ is a hyper-parameter set to $1e-3$. This hyper-parameter acts as a pseudo-Huber loss and smooths the errors smaller than $\epsilon$. 

\textcolor{black}{In a series of papers \cite{ED-DSRN, xu2021edpn,  abuolaim2021ntire}, the loss function defined either as Charbonnier  or $L1$ norm, is improved by requiring a high SSIM. This results in adding the regularization term:}  
\begin{equation}
{L}_{SSIM} = 1 - SSIM({\widehat{I}}_{i,j}^{aif},{I}_{i,j}^{aif})
\end{equation}
which makes the complete deblurring loss function to be: 
\begin{equation}
L_{deblur} = {L}_{charb}+\Psi {L}_{SSIM}
\label{deblur_loss}
\end{equation}
where $\Psi$ is a weight set to 4.

\subsubsection{2HDED loss function}

With the depth and deblurring losses defined as in eq. \ref{depth_loss} and \ref{deblur_loss}, we define the following total loss for 2HDED:NET training:

\begin{equation}
L_{2HDED} = L_{depth} + \lambda L_{deblur}  \label{total_loss}
\end{equation}

In our experiments, we tested several versions of $L_{2HDED}$: with ${L}_{depth}$ including or not ${L}_{smooth}$, with ${L}_{deblur}$ being either $L1$ norm or ${L}_{charb}$, and with or without $SSIM$ loss. 
\textcolor{black}{ We noticed during the experiments that the model performance is very sensitive to the weighting value, which is why we paid close attention to the choice of $\lambda$. Starting from the idea that both tasks should be given the same importance, we evaluated the depth and deblurring losses separately during the training, and we settled $\lambda$ in a way that they have an approximately similar contribution to the total loss. Then we fine-tuned $\lambda$ by performing a grid search and we found that $\lambda = 0.01$ is suitable for all versions of $L_{HDED}$.}

\subsection{Accuracy measures for DFD and image deblurring}
To evaluate the accuracy of the estimated depth maps and deblurred images, we use accuracy measures that have been widely reported in previous studies. \par
For the depth estimation, we compute the root mean square error (RMSE), relative absolute error (Abs. Rel.), and thresholded accuracy $\delta$ as follows:
\begin{enumerate}
\item  $RMSE=\sqrt{\frac{1}{n}\sum_{i=0}^n(\widehat{I}_{t}^{depth}-{I}^{depth})^2}$ \\
\item $Abs.Rel.=\frac{1}{n}\sum_{i=0}^n \frac{|\widehat{I}_{t}^{depth}-{I}^{depth}|}{{I}^{depth}}$\\
\item Thresholded accuracy ($\delta$) is the percentage of pixels such that: $max (\frac{\widehat{I}_{t}^{depth}}{{I}^{depth}},\frac{{I}^{depth}}{\widehat{I}_{t}^{depth}})=\delta < threshold$ \\
\end{enumerate}
To evaluate the deblurring, we resort to two well-known metrics commonly used to measure the quality of images: Peak Signal to Noise Ratio (PSNR) and SSIM.

\section{Datasets}

2HDED:NET is trained with two types of ground truth, depth maps, and AiF images. Since until recently, DFD and image deblurring have been considered separately, the existing solutions were developed around datasets dedicated to one of these applications. The lack of datasets including defocused images, corresponding depth maps, and AiF images, determined us to work on datasets for depth applications consisting of AiF images and depth ground truth and to generate defocused images by blurring the AiF images. The synthetically defocused images have been used by many recent works \cite{Anwar,Carvalho,Song} dedicated either to depth inference or image restoration.  
Thus, our choice has been the NYU-Depth V2 dataset containing indoor scenes, and the Make3D dataset with outdoor scenes. The depth range of the datasets depends on the type of sensor used to capture the depth as well as the collection method. The depth range of NYU images is $0.7$ to $10$m and of Make3D, $0$ to $80$m. 

The NYU dataset comprises $230,000$ pairs of RGB indoor images and their corresponding depth maps.
In order to speed up the experiment, the training of 2HDED:NET has been run with a smaller dataset. We used the same split as \cite{Anwar, Carvalho, saqib_smoothing} i.e., $795$ images for training and $654$ for testing. The original size of the images captured by Microsoft Kinect is $640\times480$ pixels, but they were reduced to $561\times427$ pixels in our experiments.

The Make3D dataset consists of $534$ RGB images and depth maps representing outdoor scenes. 
To train the 2HDED:NET under the same conditions as in \cite{Anwar}, 
we split the dataset similarly, i.e., into $400$ images for training and $134$ images for testing.

To avoid  overfitting, the training set has been increased by data augmentation. We adopted the data augmentation procedure addressed in \cite{Carvalho}. Since we use defocus blur as a cue, we do not apply any data augmentation process that can affect the blur information. In the first step, all the images are centered scaled. For random flips, each individual sample is flipped horizontally by $50\%$.

\subsection{Defocus Blur Simulation}
\begin{figure}[t]
\centering{\includegraphics[height=5cm, width=9cm]{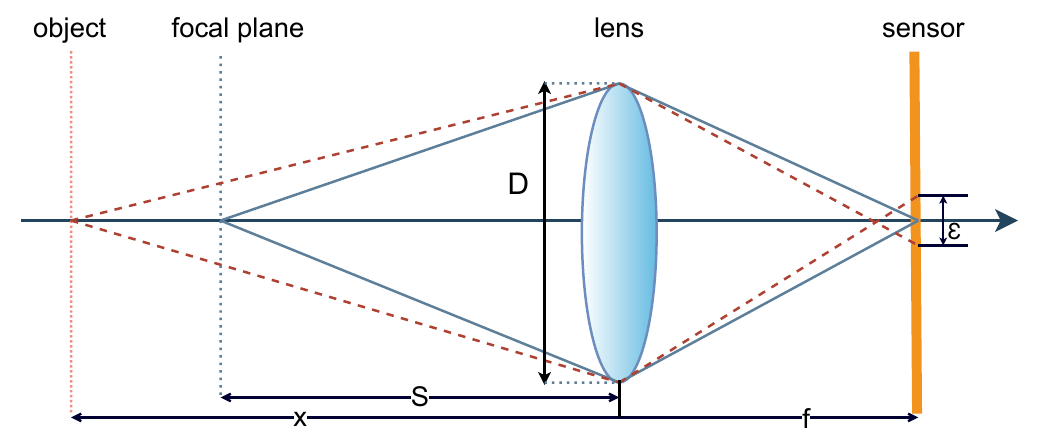}}
\caption{The thin lens model: the COC diameter $\varepsilon$ depends on the distance $x$ of the object to the lens.}
\label{thin_lens}
\end{figure}

To generate the realistic physical blur in the RGB images we adopt the procedure used by the authors in \cite{DMENET} to generate the SYNDOF dataset.
To defocus an image, they start from the thin lens model \cite{potmesil1981lens}, commonly used in computer vision (Fig.~\ref{thin_lens}). We used the same parameter values as \cite{DMENET} such as aperture size of $4.48$ cm and focal length set to $0.07m$. In Fig.~\ref{thin_lens}, $x$ is the distance to the object, $f$ is the distance from the lens to the image sensor, $D$ is the diameter of the aperture, $S$ is the distance to the in-focus plane, and $\varepsilon$ is the diameter of the circle of confusion (COC) calculated as: 
\begin{equation}
    \varepsilon = \alpha \frac{|x-S|}{x}, where \  \alpha = \frac{f}{S}D
    \label{thin_lenseq}
\end{equation}

To generate blur in the AiF image, we apply  Gaussian filters with a kernel with standard deviation  $\rho  = \varepsilon/{4}$. Similar to \cite{DMENET}, $\varepsilon$ is calculated based on the per-pixel depth values. 
As a result, we have defocused images with corresponding depth maps and AiF images.

\section{Results}
For the experimental results, we divided our analysis into the following sections:
\begin{itemize}
    \item Depth estimation and image deblurring results with various loss functions. 
    We tested simple solutions like $L_{charb}$ for deblurring and $L_1$ for depth and improved our results gradually, by adding regularizations consisting of SSIM for deblurring and smoothing for depth.
    \item Results with two heads and one head ablated to see the effectiveness of the two-head architecture.
    \item Finally, we compare our results to the state-of-the-art results for depth or image deblurring, obtained on the NYU-v2 and Make3D benchmarks. 
\end{itemize}
Our network is implemented using the PyTorch package in Python environment
The entire training session takes approximately 9 hours on an NVIDIA Quadro GV100 GPU with $32$ GB memory. We trained 2HDED:NET for $500$ epochs with a batch size of $4$ images. We use Stochastic Gradient Descent (SGD) optimizer with an initial learning rate of $0.0002$. The initial learning rate is reduced 10 times after the first $300$ epochs, this allows for large weight changes at the beginning of the learning process and small changes towards the end of the learning process. As for the total number of network parameters, our network is much lighter than \cite{Anwar}.  Specifically, our 2HDED:NET comprises a total of $41M$ parameters, whereas \cite{Anwar} employs a network with $138M$ parameters, three times higher than our model.

\begin{table*}[b]
\centering
\caption{Results on NYU-v2 dataset with different loss functions for depth estimation and image deblurring. }
\begin{tabular}[h]{|l|c|c|c|c|c|c|c|c|c|}
\hline
{\textit{Loss Function}}  &  \multicolumn{5}{|c|} {Depth Estimation} &  \multicolumn{2}{|c|} {Image Deblurring} \\
\hline
 & {\textit{RMSE}} $\downarrow$  &  {\textit{Abs. rel}} $\downarrow$ 
 & {\textit{$\delta(1)$}} $\uparrow$ & {\textit{$\delta(2)$}} $\uparrow$ & {\textit{$\delta(3)$}} $\uparrow$ & {\textit{PSNR}} $\uparrow$ & \ {\textit{SSIM}} $\uparrow$   \\

\hline
 {$L_1^{Depth}+\lambda L_{charb}^{Deblur} $} & 0.285 &  0.035 &  0.820 & 0.880 & 0.970 & 33.55 &  0.983 \\ 
 \hline
 {$L_1^{Depth}+\lambda L_{1}^{Deblur}$} & 0.292 &  0.068 & 0.799  & 0.819 & 0.891 &30.38 & 0.90  \\
 \hline
 ({$L_1^{Depth}+\mu L_{grad})+\lambda L_{charb}^{Deblur}$} & 0.244 & 0.029 & 0.901 & 0.971 & 0.989 & 32.27 &  0.918  \\ 
 \hline
 {$L_1^{Depth}+\lambda (L_{charb}^{Deblur}+\Psi (1-SSIM))$} & 0.282 &  0.031 &  0.833 & 0.895 & 0.901 & 33.85 &  0.981  \\
 \hline
{$(L_1^{Depth}+\mu L_{grad})+ \lambda(L_{charb}^{Deblur}+\Psi (1-SSIM))$} & \bf{0.241} & \bf{0.025} & \bf{0.914} & \bf{0.979} & \bf{0.995} & \bf{34.84} &  \bf{0.989}  \\ 
 \hline
 \end{tabular}
\label{tab:results_losses_terms}
\end{table*}

\begin{table*}[b]
\centering
\caption{Effect of ablating one head. Results on NYU-v2 dataset }
\vspace{2mm}
\begin{tabular}[ht]{|l|c|c|c|c|c|}
\hline
 \multicolumn{6}{|c|} {Image Debluring} \\ 
\hline
   & \multicolumn{2}{|c|} {\textit{PSNR}$\uparrow$}   & \multicolumn{3}{|c|} {\textit{SSIM}$\uparrow$}    \\
 \hline
 {With both heads } & \multicolumn{2}{|c|} {\bf{34.849}} & \multicolumn{3}{|c|} {\bf{0.989}}  \\ 
 \hline
 {Without depth head} & \multicolumn{2}{|c|} {31.941} & \multicolumn{3}{|c|}  {0.919}  \\ 
 \hline
 {Gain} & \multicolumn{2}{|c|} {2.899} & \multicolumn{3}{|c|} {0.07}  \\ 
 \hline
 \hline
 \multicolumn{6}{|c|} {Depth Estimation} \\ 
 \hline
 {\textit{}}  & {\textit{RMSE}} $\downarrow$   & \ {\textit{Abs. rel}} $\downarrow$  & \ {\textit{$\delta(1)$}} $\uparrow$  & \ {\textit{$\delta(2)$}} $\uparrow$ & \ {\textit{$\delta(3)$}} $\uparrow$ \\
 \hline
 {With both heads} &\bf{0.24} &  {\bf{0.025}} &  \bf{0.91} &  \bf{0.97} & \bf{0.99} \\ 
 \hline
 {Without deblurring head} & 0.29 & 0.075  & 0.84 & 0.89 & 0.94 \\ 
\hline
{Gain} & 0.05 & 0.05  & 0.07 & 0.08 & 0.05 \\ 
\hline
\end{tabular}
\label{tab:1Hdeblurring}
\end{table*}

\subsection{Effect of Loss Functions}

In this subsection, we perform a set of experiments consisting of training the 2HDED:NET with the various loss functions described in section \ref{lossfunctions}. 
To select among the multiple options existing for our combined depth and deblur loss function, we adopt a simple to complex approach. In the first step, we use a photometric error for deblurring and $L1$ norm for depth accuracy. For deblurring, we test with $L1$ and Charbonnier loss functions, the latter being a smoothed version of $L1$. To ensure an equal contribution of the two components, the deblurring error is weighted by $\lambda = 0.01$. We maintain the principle of equal contribution over the entire experiment.   

Table \ref{tab:results_losses_terms} presents results obtained on the NYU dataset. The depth estimation has a good accuracy even for this simple loss function. The RMSE is under $0.3$ and there are no significant differences when the photometric error for deblurring switches from Charbonnier to $L1$, where the accuracy is only slightly worse. This is not the case for deblurring, where the use of Charbonnier improves the PSNR by $3 dB$ comparing with $L1$, providing on average, a quality of $33.554 dB$ for the test set. Therefore, we choose Charbonnier for the subsequent experiments.   

\begin{figure}[t]
\centering{\includegraphics[width=1\linewidth]{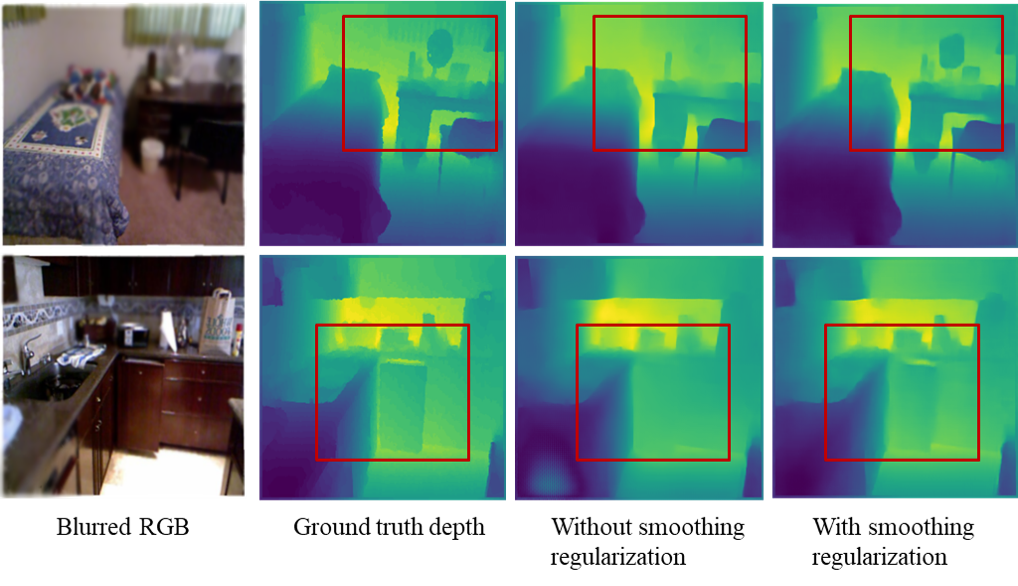}}
\caption{Results with and without the smoothing regularization for depth estimation on NYU-v2 dataset. The deblurring loss is ${L}_{charb}$. The rectangular crops illustrate areas where new details emerge when using smoothing regularization.}
\label{fig:depth_nyu_smoothing}
\end{figure}

In the second step, we alternately improve on deblur and depth losses by adding an SSIM-based term to ${L}_{charb}$ and smoothing regularization to $L1$. These additional terms are weighted by $\Psi=4$ and $\mu=0.001$, respectively. The smoothing regularization improves the RMSE depth accuracy from $0.285$ to $0.244$ on average but it worsens the deblurring results by more than $1$ dB. This apparently small difference in RMSE can impact significantly on the quality of depth maps as it can be seen from the example in Fig.~\ref{fig:depth_nyu_smoothing}, where new details are emerging when the smoothing regularization is added. 

\begin{figure}[!ht]
\centering{\includegraphics[width=1\linewidth]{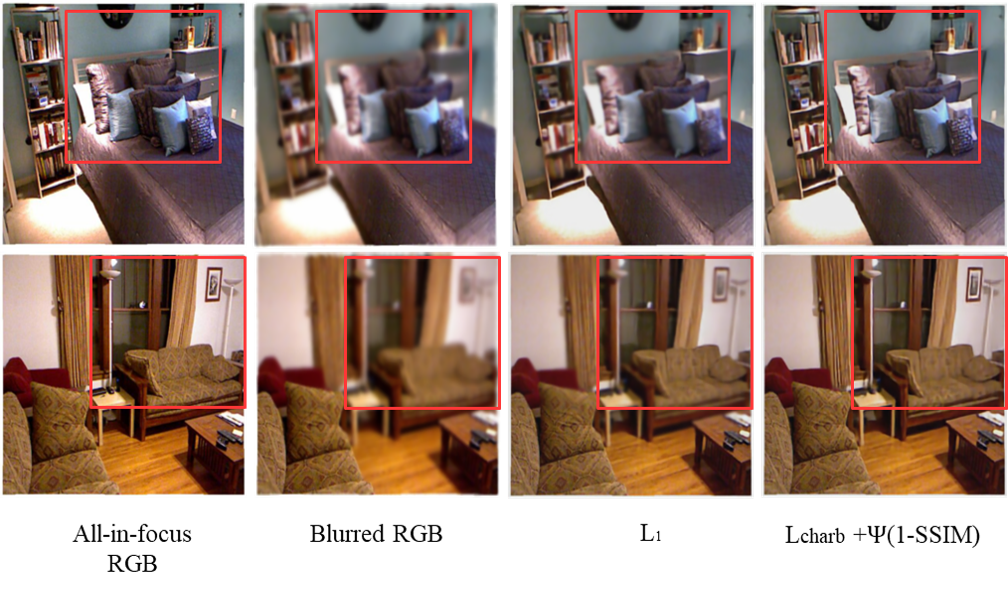}}
\caption{Deblurring results with $L_1$ and $L_{charb}+\Psi(1-SSIM)$ default loss on NYU-v2 dataset. The depth loss is $L_1$.}
\label{fig:deblurring_comparison}
\end{figure}

The introduction of SSIM term brings benefits to both depth estimation and deblurring. The RMSE decreases to $0.282$ and the PSNR becomes higher by $0.3$ dB on average.  
The deblurring results on simple $L_1$ loss and the default loss, which is $L_{charb}+\Psi(1-SSIM)$, can be seen in Fig.~\ref{fig:deblurring_comparison}.

Finally, we combine the photometric errors and the two regularizations -- SSIM and smoothing -- in a unique loss function. The network trained with this loss function achieves the best results both in depth accuracy and deblurring. The RMSE of depth touches the lowest level of $0.241$ and the PSNR of the deblurred images is almost $35$ dB. Some examples are depicted in Fig. \ref{fig:depth_nyu}
 to \ref{fig:make3d} and commented in the next subsection.

\subsection{Effect of head ablation}\label{1head_deblurring}

The 2HDED:NET is trained by using two kinds of ground truth values, the AiF image and the depth map. One is ingested by the deblurring head, the other one by the depth head. They contribute together to the training of the network as long as the loss function combines the depth and deblurring errors. Nevertheless, the network could be trained and still run by keeping a single head.  

In this subsection, we evaluate the benefits of the two-headed architecture to the overall accuracy of the model. Hence, we alternately remove one head and retrain the network by using either the AiF or depth error depending on what head is preserved. 
Table \ref{tab:1Hdeblurring} shows the results. For image deblurring, when the depth head is ablated and the loss ${L}_{charb} +\Psi (1-SSIM)$ is used, the PSNR decreases by almost $3$ dB, from $34.849$ dB to $31.941$ dB. Similarly, by ablating the deblurring head, the accuracy of the estimated depth maps becomes worse. On average, the RMSE increases by $0.05$.

Thus, it is clear that 2HDED:NET achieves the best results when both heads are used together. Each head improves the results of the other one by complementing the ground truth, even if it is of a different nature. 
\begin{table*}[b]
\centering
\caption{Comparison of 2HDED:Net with SoA methods for depth estimation and image deblurring on NYU-v2 and Make3D datasets.}
\vspace{2mm}
\begin{tabular}[ht]{|c|c||c|c|c|c|c|c||c|c|}

\hline
 \multicolumn{3}{|c|} {}  &\multicolumn{5}{|c||} {Depth Estimation} & \multicolumn{2}{|c|} {Deblurring} \\  
 \hline
 \multicolumn{10}{|c|} {NYU dataset} \\  
 \hline
 {\textit{Method}} & Depth & Deblur & {\textit{RMSE}} $\downarrow$ & {\textit{Abs. rel}} $\downarrow$ &  {\textit{$\delta(1)$}} $\uparrow$ &  {\textit{$\delta(2)$}} $\uparrow$ &  {\textit{$\delta(3)$}} $\uparrow$ &  {\textit{PSNR}} $\uparrow$ &  {\textit{SSIM}} $\uparrow$  \\
\hline
 
 {Tang et al. \cite{tang2021encoder}}  & $\checkmark$ & $\times$ &  0.579 & 0.132 & 0.826 & 0.936 & 0.992 & --  & --  \\ 
 \hline
 {Chang et al. \cite{chang}}  & $\checkmark$ & $\times$ & 0.433 & 0.087 & \textbf{0.930} & \textbf{0.990} & \textbf{0.999} & --  & --  \\ 
 \hline
 {Dong et al. \cite{mobilexnet}}  & $\checkmark$ & $\times$ & 0.537  & 0.146 & 0.799 & 0.951 & 0.988 & --  & --  \\ 
 \hline
 {Song et al. \cite{Song}}  & $\checkmark$ & $\times$ & 0.154  &  \textbf{0.028} & -- & -- & -- & --  & --  \\ 
 \hline
 {Carvalho et al. \cite{Carvalho}}  & $\checkmark$ & $\times$ & \textbf{0.144} &  0.036 & -- & -- & -- & --  & --  \\ 
 \hline
 {Gur et al. \cite{Gur}}  & $\checkmark$ & $\times$ & 0.766 &  0.255 & 0.691 & 0.880 & 0.944  & --  & --  \\
  \hline
 {Zhang et al.\cite{zhang2017beyond} (DnCNN) }  &  $\times$  & $\checkmark$ & -- &  -- & -- & -- & --  & 32.43 & 0.67  \\
  \hline
 {Zhang et al.\cite{zhang2017learning} (IrCNN) }  & $\times$  & $\checkmark$ & -- &  -- & -- & -- & --  & \textbf{35.46}   & \textbf{0.99}  \\
 \hline
 {Anwar et al. \cite{Anwar}}  & $\checkmark$ & $\checkmark$ & 0.347 &  0.094 & -- & -- & -- & 34.21 & --  \\ 
 \hline
 {2HDED:Net} & $\checkmark$ & $\checkmark$ & {0.244} & {0.029} & 0.914 & 0.979 & 0.995 & 34.85 & 0.99  \\ 
 \hline
 \hline
 \multicolumn{10}{|c|} {Make3D dataset} \\  
 \hline
  & & & \multicolumn{2}{|c|} {C1-Error} & \multicolumn{3}{|c||} {C2-Error} & &  \\  
 \hline
  & Depth & Deblur & {\textit{RMSE}} $\downarrow$  &  {\textit{Abs. rel}} $\downarrow$ & {\textit{RMSE}} $\downarrow$ & \multicolumn{2}{|c||} {\textit{Abs. rel} $\downarrow$}  & 
 {\textit{PSNR}} $\uparrow$ &  {\textit{SSIM}} $\uparrow$  \\
\hline
{Fu et al. \cite{Fu}}  & $\checkmark$ & $\times$ & \textbf{3.970} & 0.157 & 7.32 & \multicolumn{2}{|c||} {\textbf{0.162}} & --  & --  \\ 
 \hline
{Gur et al. \cite{Gur}}  & $\checkmark$ & $\times$ & 8.822 & 0.568 & 10.147 & \multicolumn{2}{|c||} {0.575} & --  & --  \\ 
\hline
{Zhang et al.\cite{zhang2017beyond} (DnCNN)}  & $\times$ &  $\checkmark$ & -- & -- & -- & \multicolumn{2}{|c||} {--} & 23.16  & 0.68  \\
\hline
{Zhang et al.\cite{zhang2017learning} (IrCNN)}  & $\times$ & $\checkmark$ & -- & -- & -- & \multicolumn{2}{|c||} {--} & \textbf{27.09}  & 0.70  \\
\hline
{2HDED:Net}  & $\checkmark$ & $\checkmark$ & 4.178 & \textbf{0.153} &  \textbf{6.132} & \multicolumn{2}{|c||} {0.170} & 24.76  & \textbf{0.78}  \\ 
\hline
\end{tabular}
\label{tab:DD_comparison_SOA}
\end{table*}

\subsection{Comparison with SoA methods for Depth Estimation and Image Deblurring}

We compare 2HDED:NET with some state-of-the-art solutions based on neural networks for depth estimation and image deblurring. Since in the literature, there are very few networks that solve simultaneously the problems of DFD and image deblurring \cite{Anwar,Gur}, we also consider recent methods dedicated exclusively to depth estimation. The blur is rarely taken into consideration in such cases \cite{tang2021encoder, chang, mobilexnet}, most of the networks being trained on AiF images. 
Table \ref{tab:DD_comparison_SOA} presents in the left half, results for depth estimation obtained with networks trained on NYU and Make3D dataset. 

For NYU dataset, the best accuracy in terms of RMSE is obtained by Carvalho et al.  \cite{Carvalho} and Song et al. \cite{Song}, both trained on defocused images with the sole purpose of generating depth maps. Their performances are very close, \cite{Song} outperforms \cite{Carvalho} on $Abs. rel$ but not on RMSE. The approach proposed in \cite{Song} demonstrates improvement in performance by utilizing pairs of images with varying degrees of defocus to estimate depth, thereby providing additional ground truth information.

From the same category of networks using defocused i-mages, there are \cite{Gur} and \cite{Anwar}. They are the most representative for our comparison since these networks handle both depth maps and blurred images. On average, the depth maps accuracy of \cite{Anwar} is worse by $0.2$ in RMSE comparing with the best result in \cite{Carvalho}. Gur et al. \cite{Gur} lags behind with an RMSE of $0.766$ but the results are still remarkable given the fact that they use self-supervised learning. 
2HDED:NET is at half way between \cite{Carvalho} and \cite{Anwar} with a RMSE of $0.241$. In the category of networks handling both depth and deblurred images, our 2HDED:NET is the best in all metrics.

We also present results for three recent networks trained on AiF images to generate exclusively depth maps. The average RMSE ranges between $0.433$ and $0.579$, well inferior to the results of \cite{Carvalho} or \cite{Song} and to our result. This difference proves the effectiveness of the defocus in the training set. The defocus is an additional source of information, independent of the scene geometry, which is commonly exploited by neural networks.
\begin{figure}[t]
\centering{\includegraphics[width=\linewidth]{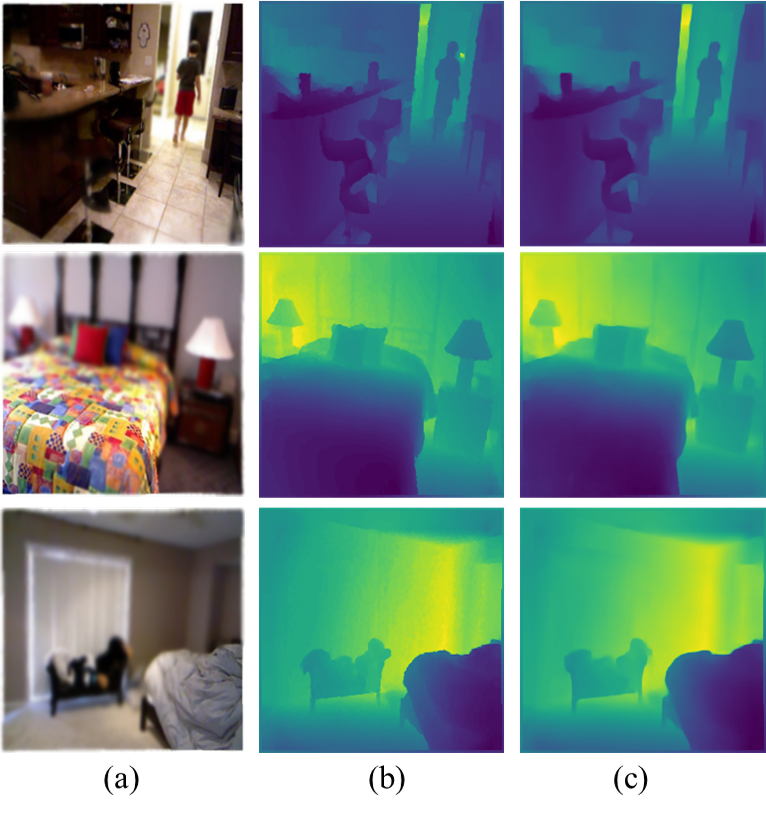}}
\caption{2HDED:NET results for depth estimation on NYU-v2 dataset using the default loss $L_{2HDED}$: (a) RGB defocused image (b) Depth ground truth (c) Estimated Depth.}
\label{fig:depth_nyu}
\end{figure}
\begin{figure}[t]
\centering{\includegraphics[width=\linewidth]{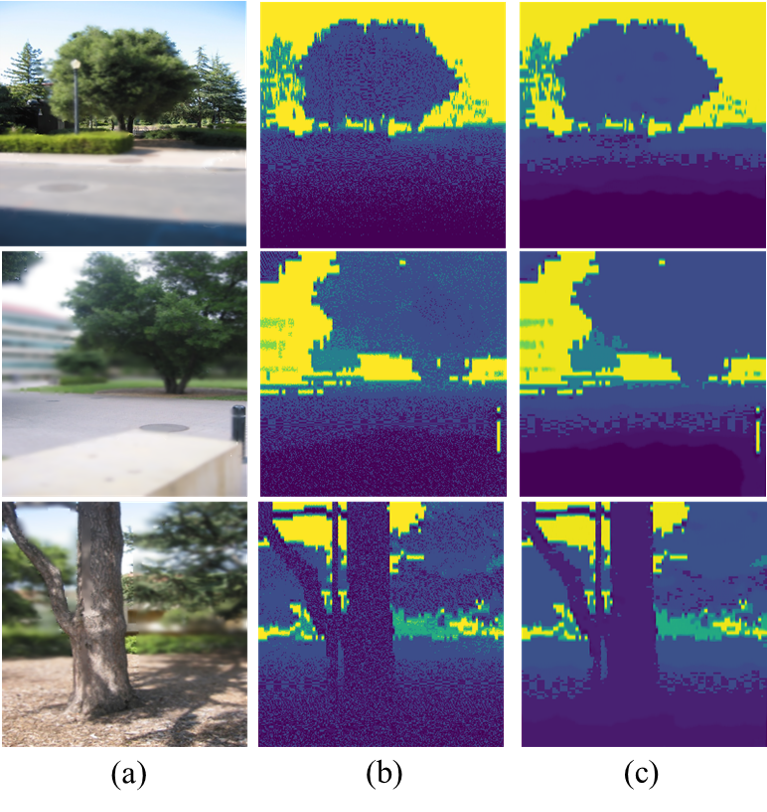}}
\caption{2HDED:NET results for depth estimation on Make3D dataset using the default loss $L_{2HDED}$: (a) RGB defocused image (b) Depth ground truth (c) Estimated depth.}	
\label{fig:depth_3d}
\end{figure}
\begin{figure*}[ht]
	\centering
	\begin{minipage}{1\columnwidth}
		\centering
		\includegraphics[width=0.95\linewidth]{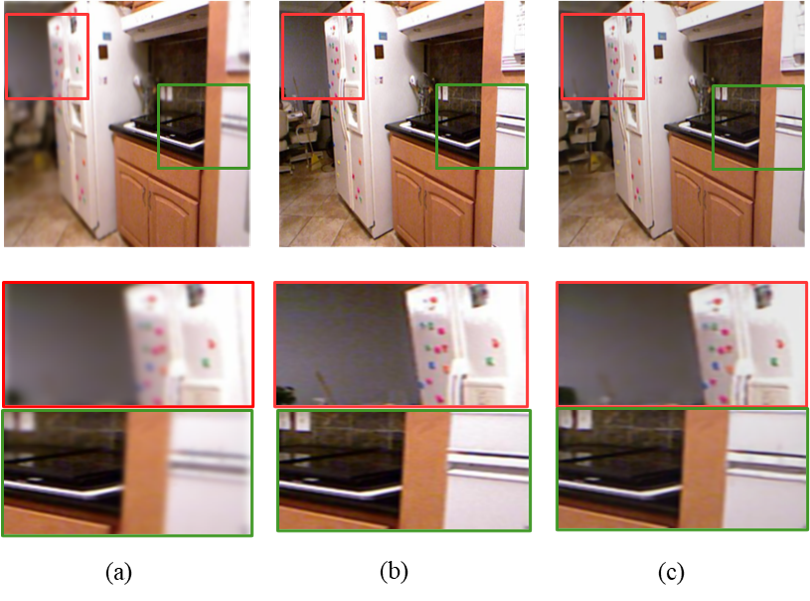}
		\label{label1}
	\end{minipage}
	\begin{minipage}{1\columnwidth}
		\centering
		\includegraphics[width=0.95\linewidth]{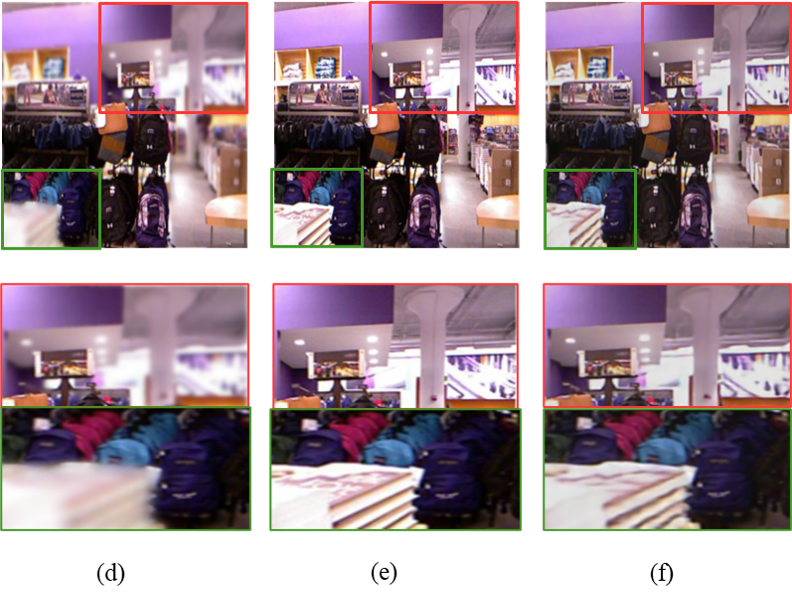}
		\label{label2}
	\end{minipage}
	\begin{minipage}{1\columnwidth}
		\centering
		\includegraphics[width=0.95\linewidth]{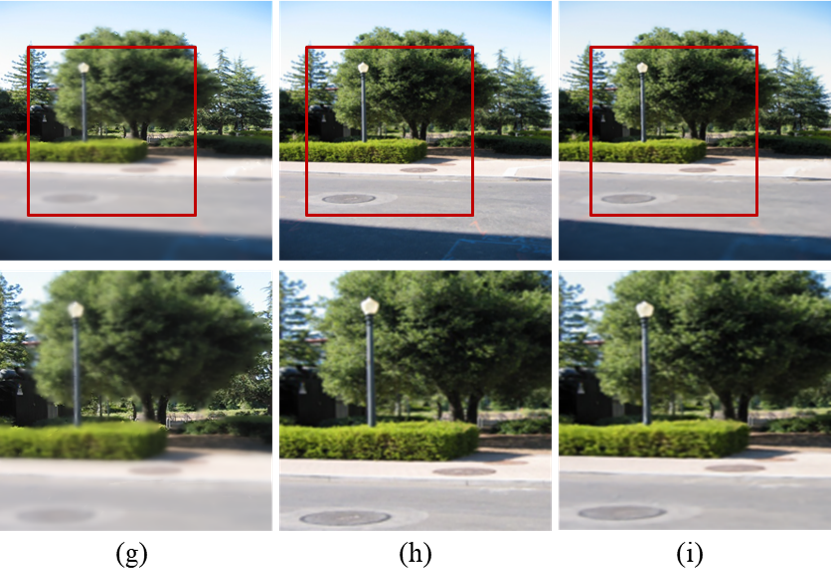}
		\label{label3}
	\end{minipage}
	\begin{minipage}{1\columnwidth}
		\centering
		\includegraphics[width=0.95\linewidth]{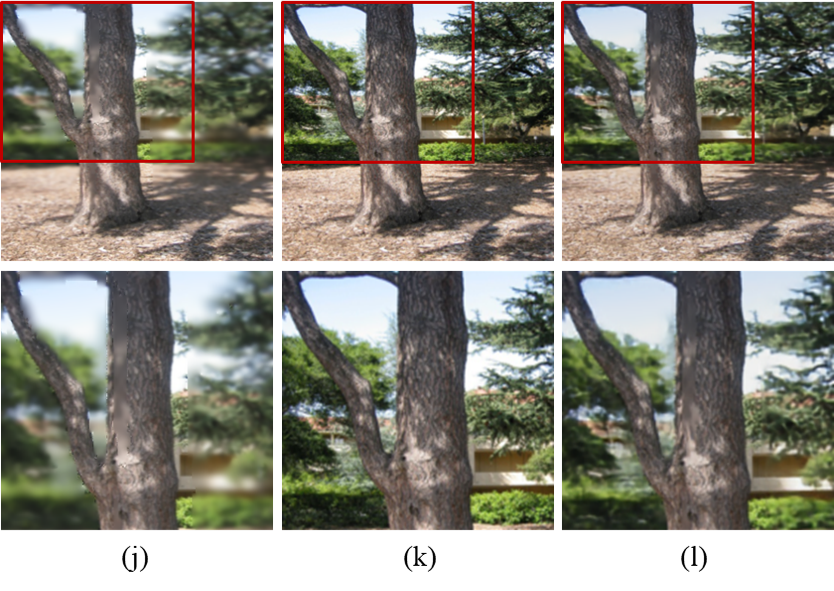}
		\label{label4}
	\end{minipage}
\caption{2HDED:NET results for deblurring on NYU-v2 and Make3D dataset. From left to right: (a) defocused image, (b) ground truth AiF image, and (c) deblurred image. Similarly, (d), (e), and (f) for a different scene. Zoomed-in patches are shown below each scene.}	\label{fig:make3d}
\end{figure*}
Figure ~\ref{fig:depth_nyu} depicts three examples of depth  maps obtained with 2HDED:NET. The visual comparison with the ground truth shows high-quality results. 
The smoothing regularization added to $L1$ loss instructs the network to produce depth maps with sharp edges and smooth, homogeneous regions that match well the ground truth and have significantly fewer artifacts. 

We also evaluate the performance of 2HDED:NET on Make3D dataset, which consists of outdoor scenes. 
The common approach to measuring the depth accuracy on this data set is to estimate errors on two depth ranges: C1 for depth up to $70$m and C2 up to $80$m \cite{c1c2}. The results for Make3D dataset are shown in the lower half of Table \ref{tab:DD_comparison_SOA}. 
Earlier methods \cite{Fu, Gur}, trained on Make3D dataset, are also displayed for comparison. 
Our analysis shows that 2HDED:NET performs better than \cite{Gur} in all metrics and for both ranges. 
In what concerns \cite{Fu}, our results are better in terms of $Abs. rel.$ on C1 range and $RMSE$
on C2.

Figure~\ref{fig:depth_3d} depicts qualitative results for three different scenes. 2HDED:NET manages to correctly extract the depth of both near and distant regions.

\begin{figure*}[t]
		\centering
		\includegraphics[width=5.5 in]{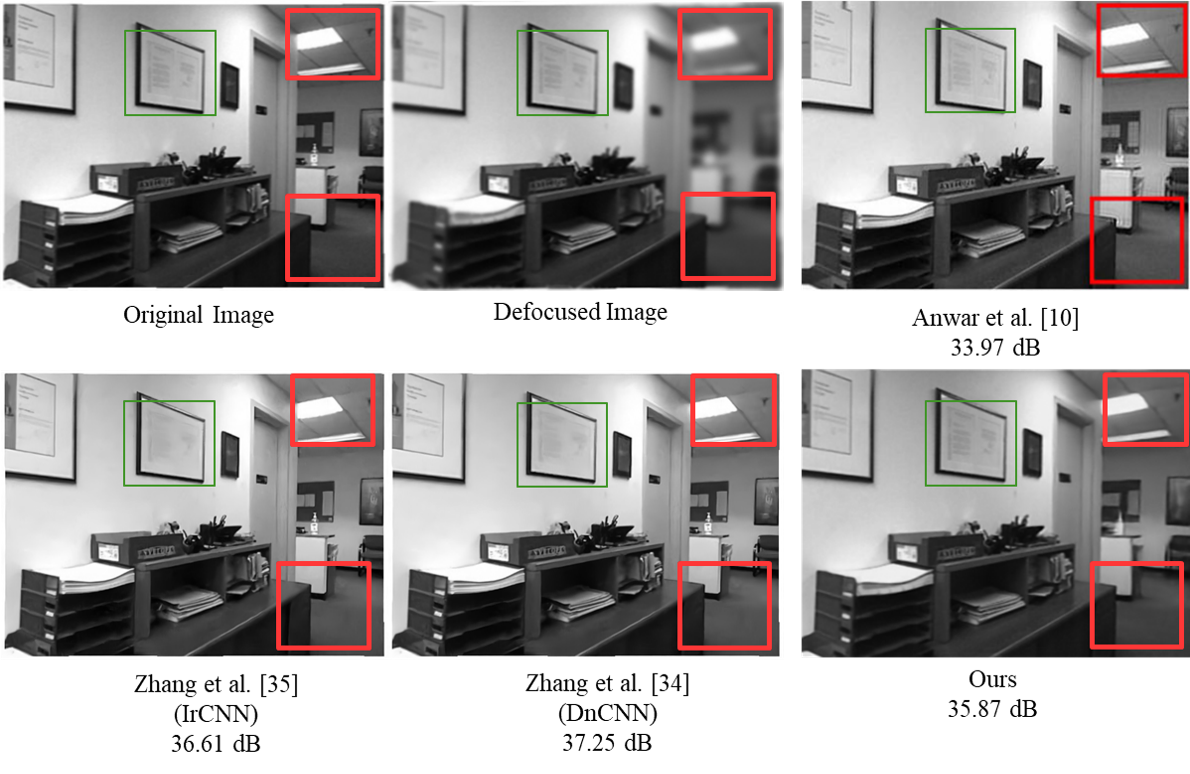}
		\label{label5}
	\vspace{-0.4cm}
	\hspace*{25em}
\caption{Comparison of 2HDED:NET with the pipeline solution of Anwar et al. \cite{Anwar}, and two general methods for image restoration \cite{zhang2017beyond, zhang2017learning}: an example from NYU-v2 dataset.}	\label{fig:NYU}
\end{figure*}
\begin{figure*}[!t]
		\centering
		\includegraphics[width=5 in]{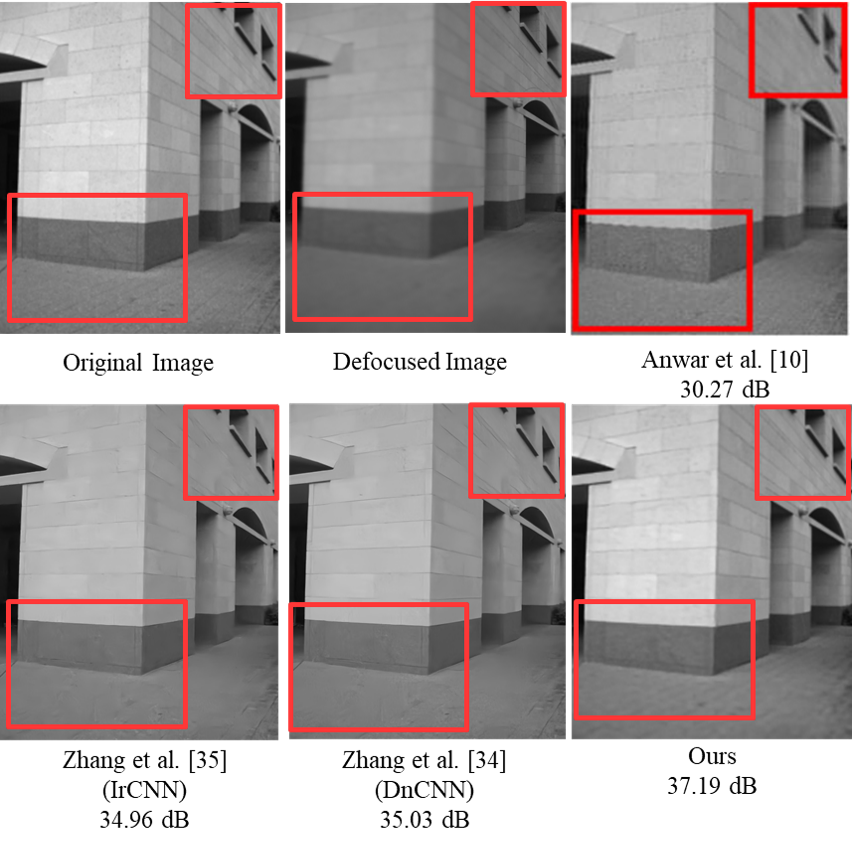}
	\vspace{-0.5cm}
\caption{Comparison of 2HDED:NET with the pipeline solution of Anwar et al. \cite{Anwar}, and two general methods for image restoration \cite{zhang2017beyond, zhang2017learning}: an example from Make3D dataset.}	\label{fig:make3dSOA}
\end{figure*}

Regarding image deblurring, in order to have a fair comparison, we selected only methods that are tested on NYU and/or Make3D benchmarks. The number of such methods is limited as these benchmarks are typically employed for depth estimation, rather than deblurring. 
Thus, for the NYU dataset, we selected \cite{Gur, qiu2019world, Anwar} as a baseline, and for the Make3D dataset, only \cite{Anwar}. The results are shown in the right half of Table \ref{tab:DD_comparison_SOA}.

2HDED:NET performs better than the selected methods for NYU dataset, where it achieves an average PSNR of $34.85$ dB. Comparing with \cite{Anwar}, which is the main competitor, our results are superior by a margin of $0.64$ dB. 
2HDED:NET outperforms this method also on Make3D dataset, where the average PSNR is higher by $3$db. 

Some qualitative results are depicted in Fig.~\ref{fig:make3d}. On the first row, there are two scenes from the NYU dataset. For each scene, the AiF image \ref{fig:make3d}(b), its artificially defocused counterpart \ref{fig:make3d}(a), and the deblurred version output by 2HDED:NET \ref{fig:make3d}(c) can be compared. It can be seen how the blur is reduced both from near and far ranges. Two small areas, one with the stickers on the fridge in the foreground and the other with the margin of the hob closer to the camera are zoomed in on the second row. Fig.~\ref{fig:make3d}(c) shows how the details come to the surface after deblurring in both  distant and near areas. 2HDED:NET works similarly on the second scene depicted in Fig.~\ref{fig:make3d}(d--f). Far and near-range patches are zoomed in order to prove the quality of the restored details.    

For the outdoor scenes in the Make3D dataset, the quality of the restored image can be observed from the two examples in Fig.~\ref{fig:make3d}(g-l). The street light and the manhole cover that appear highly defocused in Fig.~\ref{fig:make3d}(g) are well restored in Fig.~\ref{fig:make3d}(i). Similarly, the tree branches in the second scene of Fig.~\ref{fig:make3d}(j) are obviously restored by 2HDED:NET in Fig.~\ref{fig:make3d}(l).

{\color{black}We also compared our results with the two networks, IrCNN and DnCNN, proposed by Zhang Kai et al. in \cite{zhang2017beyond} and \cite{zhang2017learning} for deblurring. Since the reported results were for other benchmarks, we retrained and tested the networks on our datasets. The DnCNN is under 2HDED:NET in PSNR and SSIM of the deblurred images, while IrDNN overcomes our solution by 0.6 dB on average on the NYU dataset, and by 2.33 dB on Make3D. Still, there are cases like the image in Fig.~\ref{fig:make3dSOA} with fine textures, where our network performs much better (a PSNR gain of 2 dB). It seems that the fine textures are better restored by 2HDED:NET.}   

In the evaluation of 2HDED:NET, we placed particular emphasis on comparing it with the network proposed by Anwar et al. \cite{Anwar}. Although both networks provide depth maps and deblurred images, Anwar's network utilizes a pipeline processing approach, making it a suitable point of comparison with our network.
 
In Fig.~\ref{fig:NYU}, we give an example of an image from the NYU dataset restored by both \cite{Anwar} and 2HDED:NET. Our method achieves a PSNR of $35.87$ dB, which is almost $2$ dB higher than that of \cite{Anwar} for the same image. The blur removal can be well observed in the areas delimited by the red rectangles: the light on the ceiling and the edges of the furniture.
Another example, from the outdoor Make3D dataset, is depicted in Fig.~\ref{fig:make3dSOA}. In this particular image, 2HDED:NET archives a PSNR of $37.19$ dB, which is higher by almost $7$ dB when compared to that obtained by \cite{Anwar}. The highly textured area of the wall, with tiles that are almost invisible in the image obtained by \cite{Anwar}, is well restored in the image output by 2HDED:NET. 
Another advantage of 2HDED:NET with respect to \cite{Anwar} is the speed of computation. Once trained, 2HDED:NET generates the restored images very quickly, while in the case of \cite{Anwar}, the restoration is a long process because of pixel-wise non-blind deconvolution.

\section{Conclusion and future work}
\label{sec:conclusion}

In this work, we presented a novel deep convolutional neural network that estimates depth and restores the AiF images from a single out-of-focus image. 
The proposed network has a two-headed architecture consisting of an encoder and two parallel decoders, each of which with different roles: one outputs the depth map and the other the deblurred image. 
The formulation of an architecture that estimates the depth maps while removing blur from out-of-focus images, distinguishes our network from existing methods that are using pipeline processing. 
By parallelizing the tasks, the complexity of the network is reduced, while the depth estimation and blur removal work together toward performances that prove to be superior or close to the state-of-the-art results. 
Extensive tests on indoor and outdoor benchmarks have shown that 2HDED:NET outperforms the existing pipeline networks in both DFD and image deblurring. 
For the novel architecture of 2HDED:NET, we have proposed a new loss function that fuses depth and AiF errors, traditionally used separately in deep learning.

Since we experimented with synthetically blurred datasets, our future work will focus on developing a real defocused dataset containing depth ground truth, AiF and naturally defocused images. 

\bibliographystyle{IEEEtran}
\bibliography{main.bib}

\vspace{-10pt}
\vskip 0pt plus -1.1fil
\begin{IEEEbiography}[{\includegraphics[width=1in,height=1.25in,clip,keepaspectratio]{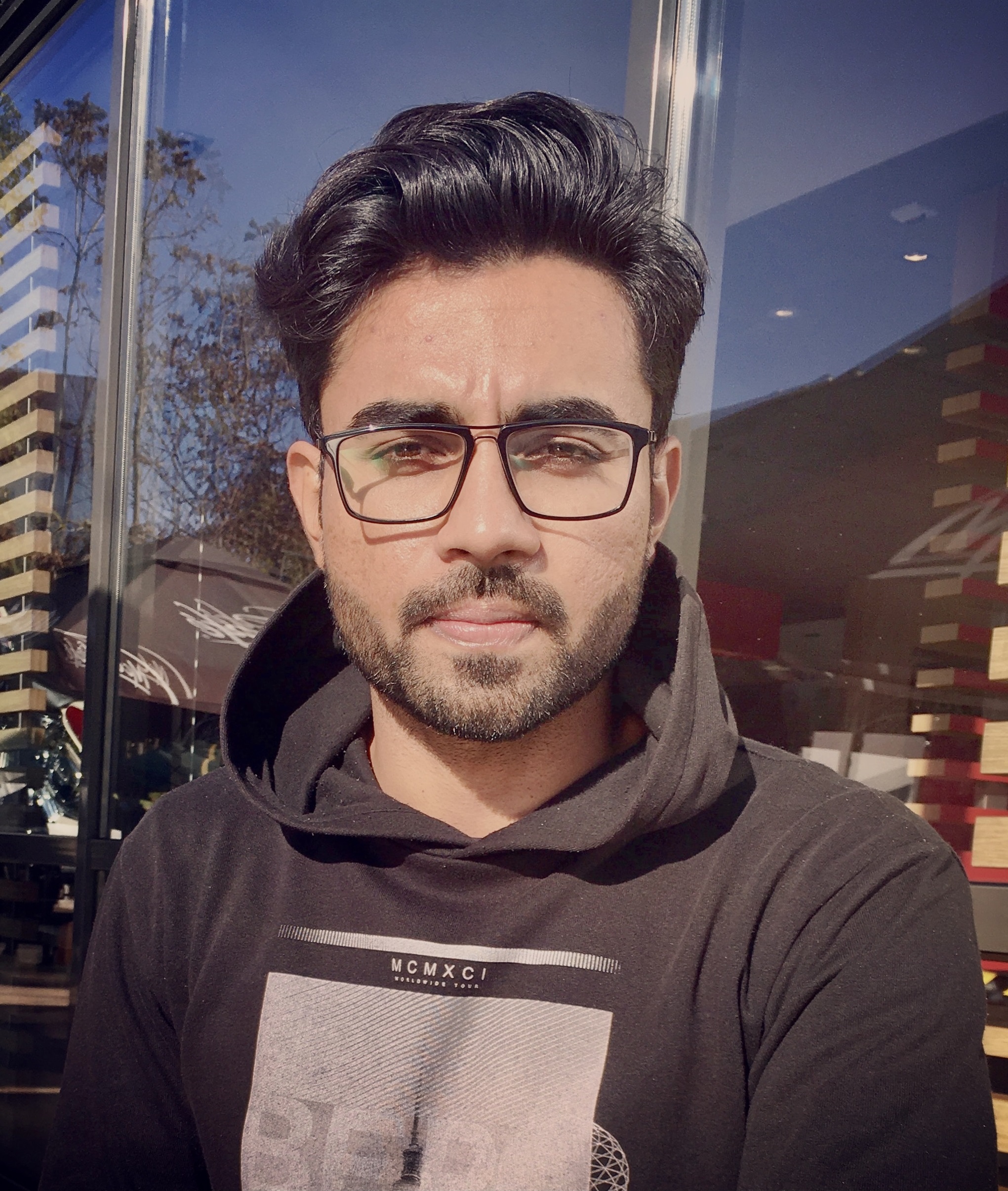}}]{Saqib Nazir} is a Ph.D. student at CEOSpace Tech. of Polytechnic University of Bucharest (UPB), Romania. He received his Master's and Bachelors’s degrees in Computer Science from COMSATS University Islamabad, Pakistan. His role as an Early-Stage Researcher in the MENELOAS-NT Project is to acquire depth from defocus images using state-of-the-art deep learning methods. His areas of interest include "Computer Vision, Image analysis, and Machine \& Deep Learning".

\end{IEEEbiography}
\vspace{-1cm}
\vskip 0pt plus -1.1fil

\begin{IEEEbiography}[{\includegraphics[width=1in,height=1.25in,clip,keepaspectratio]{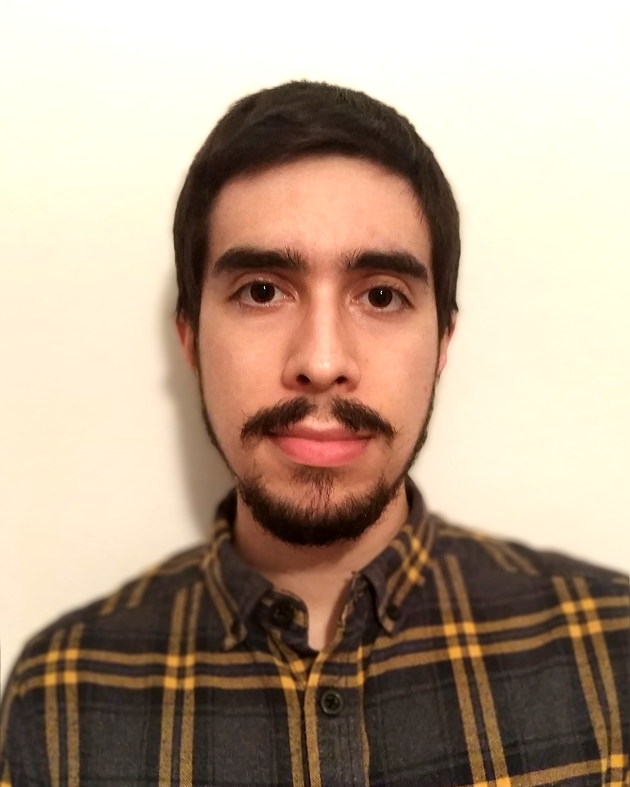}}]{Lorenzo Vaquero} is a Ph.D. student at the CiTIUS of the University of Santiago de Compostela, Spain.
He received the B.S. degree in Computer Science in 2018 and the M.S. degree in Big Data in 2019.
His research interests are visual object tracking and deep learning for autonomous vehicles.
\end{IEEEbiography}
\vspace{-1cm}
\vskip 0pt plus -1.1fil

\begin{IEEEbiography}[{\includegraphics[width=1in,height=1.25in,clip,keepaspectratio]{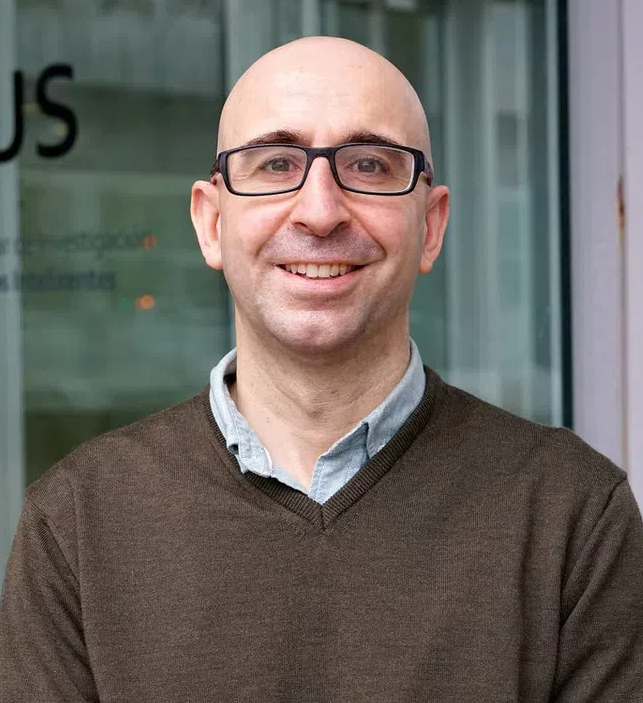}}]{V\'ictor M. Brea} is an Associate Professor at CiTIUS, University of Santiago de Compostela, Spain.
His main research interest lies in Computer Vision, both on deep learning algorithms, and on the design of efficient architectures and CMOS solutions.
He has authored more than 100 scientific papers in these fields of research.
\end{IEEEbiography}
\vspace{1cm}
\vskip 0pt plus -1.1fil
\begin{IEEEbiography}[{\includegraphics[width=1in,height=1.25in,clip,keepaspectratio]{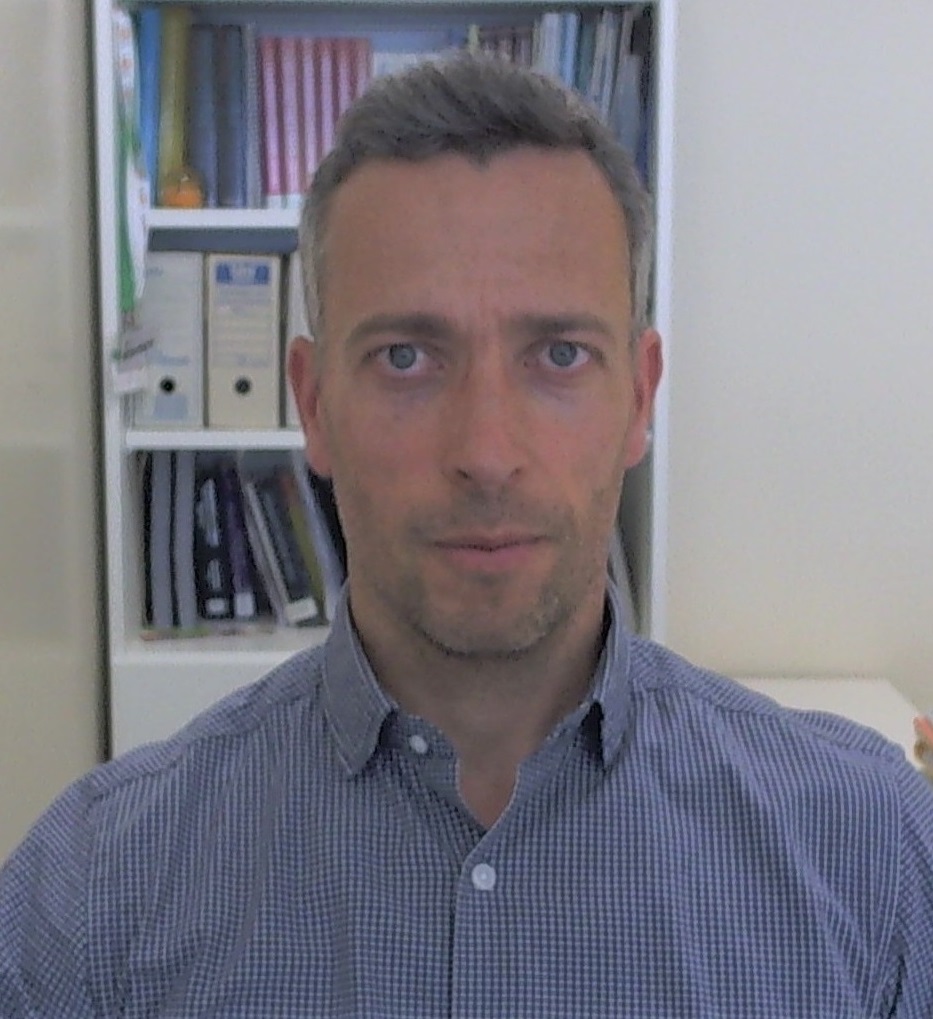}}]{Manuel Mucientes} is an Associate Professor at the CiTIUS of the University of Santiago de Compostela, Spain.
His main research interest is artificial intelligence applied to the following areas: computer vision for object detection and tracking; machine learning; process mining.
He has authored more than 100 scientific papers in these fields of research.
\end{IEEEbiography}
\vspace{1cm}
\vskip 0pt plus -1.1fil
\begin{IEEEbiography}[{\includegraphics[width=1in,height=1.25in,clip,keepaspectratio]{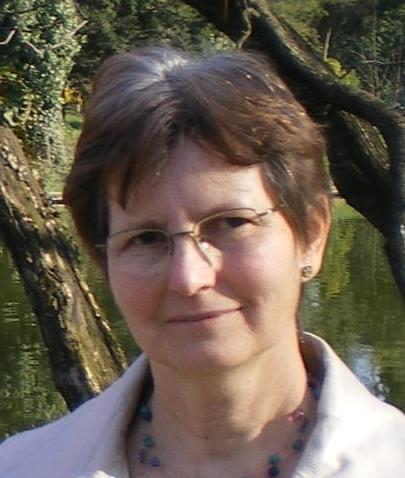}}]{Daniela COLŢUC} received MSc, PhD degrees and habilitation in Electronics, Telecommunications and Information Technology from Univ. POLITEHNICA of Bucharest, Romania. She is currently full professor with this university. She has served also as invited professor at Univ. Jean Monet in St. Etienne and Univ. de Lyon, France.
She has a sound background in Information theory with applications in image processing. In the recent years, she has worked in computational imaging.
\end{IEEEbiography}

\end{document}